\documentclass[runningheads]{llncs}
\usepackage{comment}
\usepackage{graphicx}
\usepackage{color}
\usepackage{url}
\usepackage{subcaption} 
\usepackage{multibib}
\usepackage[leqno]{amsmath}
\usepackage{examples}
\usepackage{multirow}
\usepackage{xcolor}
\usepackage{nicematrix}

\usepackage[T1]{fontenc}
\usepackage{lmodern}

\usepackage[subtle]{savetrees}

\iffalse
\newcommand{\RB}[1]{\textcolor{blue}{[RB: #1]}}
\newcommand{\EK}[1]{\textcolor{red}{[EK: #1]}}
\newcommand{\IS}[1]{\textcolor{magenta}{[IS: #1]}}

\else
\newcommand{\RB}[1]{\textcolor{blue}{\ignorespaces}}
\newcommand{\EK}[1]{\textcolor{red}{\ignorespaces}}
\newcommand{\IS}[1]{\textcolor{magenta}{\ignorespaces}}
\fi

\begin{document}

%
\title{Raising Student Completion Rates with \\ Adaptive Curriculum and Contextual Bandits}
\titlerunning{Raising Completion Rates with Bandits}
%
\author{Robert Belfer\inst{1} \and Ekaterina Kochmar\inst{1,2}  \and Iulian Vlad Serban\inst{1}
}

%
\authorrunning{Belfer et al.}

\institute{
Korbit Technologies Inc.
\and University of Bath \\
\email{robert [AT] korbit.ai}}




%

%
\maketitle
%
\begin{abstract}
We present an adaptive learning Intelligent Tutoring System, which uses model-based reinforcement learning in the form of contextual bandits to assign learning activities to students. The model is trained on the trajectories of thousands of students in order to maximize their exercise completion rates and continues to learn online, automatically adjusting itself to new activities. A randomized controlled trial with students shows that our model leads to superior completion rates and significantly improved student engagement when compared to other approaches. Our approach is fully-automated unlocking new opportunities for learning experience personalization.



\end{abstract}

%
%
%

\section{Introduction}

Intelligent Tutoring Systems (ITS) aim to provide personalized tutoring in a computer-based environment and are capable of selecting problems on an individual basis~\cite{nye2014autotutor}.
Many ITS consider the development of personalized curricula: a recommended sequence of learning activities adapted in real-time to the needs of each individual student 
~\cite{albacete2019impact,rus2014deeptutor}.
Investigating novel methods for developing personalized curricula that can adapt to millions of students and thousands of courses or domains in real-time is key to further improvements in the learning experience of students interacting with ITS.

We present {\tt Korbit}, an adaptive learning ITS leveraging reinforcement learning (RL) in order to automatically assign learning activities to students.
{\tt Korbit} is an online learning platform, where students follow a blended-learning framework combining problem-solving activities, lecture videos, Socratic tutoring and project-based learning~\cite{serban2020large}. We focus on ordering the text-based problem-solving activities, which students can answer in free-form text or as multiple-choice questions (MCQs).
If the student answers correctly, they move on to a different exercise; otherwise, they are given feedback and may try again or skip the activity and move on.
The ordering of exercises and all other activities within the same continuous topic (called \textit{learning unit}) is determined by a model-based RL system employing the LinUCB algorithm~\cite{Li2010}. 

The main contributions of this paper are two-fold: (1) we present the design and implementation of a model-based RL system for ordering learning activities based on the LinUCB algorithm; (2) we evaluate this model in a randomized controlled trial and show that it attains superior completion rates and improved student engagement when compared to alternative approaches.

\section{Background}


Personalization is key to effective learning~\cite{anania1983influence,bloom19842}. In computer-based learning environments (CBLEs),  ITS have been shown to dramatically improve student learning outcomes and engagement~\cite{kulik2016effectiveness,VanLehn} due to their ability to address individual needs and develop personalized feedback~\cite{albacete2019impact,piano2018,mu2018combining}. One of the most powerful families of algorithms deployed in CBLEs are RL algorithms,which have been successfully applied to personalize the curriculum and learning activities~\cite{doroudi2019s,Lan2016,Lopes2013} and to assess different educational interventions through the use of multi-armed bandits \cite{Liu2014,Whitehill2017,Williams2018}.

In multi-armed bandit problems, an agent sequentially selects an action and observes a reward from it, with the ultimate goal of maximizing cumulative reward over the long term.
Since actions taken by the agent at any particular time may be suboptimal,  a mix between {\em exploration} (trying out new strategies) and {\em exploitation} (picking the action deemed optimal at the time) is required in practice to maximize observed long-term reward. 
Agents are evaluated using \textit{regret}, which is defined as the cumulative expected difference between the rewards of the optimal action and the selected actions.

The {\em contextual bandit} model presents an agent with information about the current context that it can use to inform its decision.  The \texttt{LinUCB} algorithm \cite{Li2010}, which we use in this work, achieves the theoretical regret bound of $ O(\sqrt{T})$ (where $T$ is the number of timesteps), while being relatively easy to implement and less prone to numerical instability issues throughout its runtime than alternatives~\cite{Auer2003_2}. At each timestep $t=1,..., T$, the \texttt{LinUCB} agent observes the current user $u_t$, a set $A_t$ of actions, and a feature vector $\mathbf{x}_{t,a}$ for each $a \in A_t$. Each feature vector contains information about both the user $u_t$ and its corresponding action $a$, and is referred to as a context. The algorithm then computes a score $p_{t,a}$ for each action, based on its expected reward and uncertainty determined by the context vectors and its internal parameters. It receives a reward $r_{t}$ and uses it to update its internal parameters, thus improving its selection strategy.

The approach proposed here combines the {\tt LinUCB} algorithm with model-based RL, where an internal model of the environment is learned by the RL agent. By learning an internal environment model the agent may be able to reduce the amount of trial-and-error learning and better generalize across states and actions. In particular, the internal environment model may be learned from historical data, if such is available.  Several researchers have also investigated the application of model-based reinforcement learning for ITS, including learning effective pedagogical policies, selecting effective instructional sequences and personalizing curricula for students \cite{chi2011empirically,doroudi2019s,rowe2015improving}.

\section{Methodology}

We train a model that can predict a student's performance on an exercise and then use it to simulate student trajectories to pre-train the {\tt LinUCB} exercise selection model.


\noindent{\bf Dataset}:
We first extract all previous solution attempts across all $1,977$ students that created their accounts between November 2020 and July 2021 and that have attempted at least one exercise.
We retrieve $129,000$ exercise attempts across $971$ unique exercises and $61$ learning units. 
The majority of students on the platform at the time were free users, so we separate the free users and the customers in further experiments.

%

\noindent{\bf Exercise Affinity Model}:
The five possible outcomes for an interaction between a student and an exercise on our platform are defined as follows:

\vspace{-0.5em}
\begin{itemize}
\item[$\bullet$] {\bf Instant success}: The student solved the exercise correctly on the first try.
\item[$\bullet$] {\bf Eventual success}: It took the student multiple attempts to get to a correct solution. 
\item[$\bullet$] {\bf Eventual failure}: The exercise was attempted unsuccessfully until a solution was provided to the student.
\item[$\bullet$] {\bf Instant skip}: The exercise was skipped without any attempt. 
\item[$\bullet$] {\bf Eventual skip}: The exercise was attempted but eventually skipped.
\end{itemize}
\vspace{-0.5em}

\begin{figure}[h!]
\centering
    \vspace*{-0.4cm}
    \includegraphics[scale = 0.40]{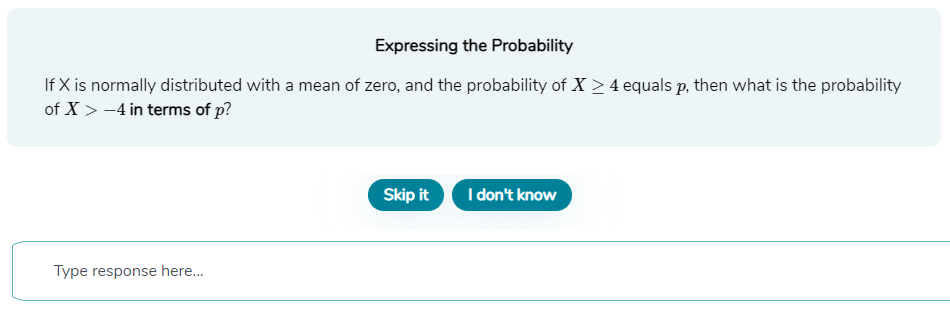}
    \caption{Example of a free-form question on our platform.}
    \label{fig:freehand}
    \vspace*{-0.4cm}
\end{figure}

First, we build a logistic regression model that uses exercise features and students' performance on previous exercises to predict the outcome on future exercises. This model will act as the ``world model" in the context of model-based RL, and will provide the agent with the outcomes when it offers an exercise to a student. We train this model by first extracting a student's exercise attempt history, which contains all of a student's attempts at solving the exercises they were presented with (both successful and unsuccessful). We then mask out an attempt on an exercise, and have the model predict the outcome of the student's attempt on this exercise.  The model's input features relate to the {\em student behavior and skills} (including the student's performance on the previous exercise in the learning unit, the student's skip rate in the learning unit, and whether or not the student has watched the video that covers the learning unit),  {\em exercise difficulty} (the historic success rate across all students on the exercise), and the {\em exercise type} (a one-hot encoding of the expected solution form and the context in which the exercise could be applied). We show an example of a free-form question in Figure \ref{fig:freehand}.

We train the model on approximately $129,000$ examples and evaluate it using $5$-fold cross-validation, observing an accuracy of $66\%$.  A baseline model that selects the majority class $100\%$ of the time achieves an accuracy of $60\%$. Although the prediction model can be refined and improved, we believe that it is good enough to be used in the context of pre-training a bandit model. We then use this model to predict students' performance on all unattempted exercises on topics that they have started. This prediction takes the form of a probability vector across the $5$ possible outcomes. A total of $165,000$ exercise attempts are predicted. These predictions allow us to simulate what would happen if the student receives an exercise that we have no record of them attempting. To train the bandit model, we draw samples from the predictions.

\noindent{\bf Bandit Model}:
Using the dataset of student trajectories and attempt predictions, we train a bandit model with {\tt LinUCB}. At each timestep, it selects which exercise to present to the student and whether it should be a MCQ or a free-form question. For each action, we compute a feature vector that encodes information about both the student and the exercise using the same features as the exercise prediction model. To simulate students' progressing through various learning units, for each unit a student has started we let the model sequentially select exercises to present to them. We define our reward function such that more desirable outcomes receive higher rewards, with the ultimate goal of maximizing average success rate for the students in the dataset, while restricting the frequency of MCQs. While doing a grid search, we observed that rewarding instant success higher than eventual success led to a higher average completion rate on the dataset. Our final reward function is as follows: $1.5$ for instant success,  $1$ for eventual success,  $0.5$ for eventual failure, and $0$ for instant and eventual skips. 
To discourage the model from always presenting an MCQ, we penalize the observed reward for MCQs by reducing it by $0.4$. Since students are more likely to correctly answer such questions, the model is more likely to observe a positive outcome when presenting them. However, the free-form questions lead to higher learning outcomes and engagement in students.
For a given $(student, \: exercise)$ pair, we use the observed outcome if the student has attempted that exercise in our dataset. Otherwise, we sample an outcome from the probabilities computed by our prediction model.

We compare overall success rates of $3$ policies on our dataset:
\vspace{-0.5em}
\begin{itemize}
\item[$\bullet$] {\sc Random} uniformly selects a new non-MCQ exercise from the current topic.
 \item[$\bullet$]  {\sc Heuristic} sorts available exercises from easiest to hardest, offers a ``medium" difficulty one at the start, decreases difficulty upon skip or failure and increases it upon success. If a student fails multiple times in a row, it begins to offer MCQs.
\item[$\bullet$]  {\sc LinUCB}: this policy is learned by our {\tt LinUCB} model.
\end{itemize}
\vspace{-0.5em}

For each policy, we simulate every student attempting the exercise presented by the policy, and keep track of the average success rate. Due to randomness, we do this $20$ times. We observe an average success rate of $58\%$ for the {\sc Random} policy, $60\%$ for the {\sc Heuristic} policy, and $64\%$ for {\sc LinUCB}. These values are consistent throughout each run, deviating by no more than $0.5\%$. Both the {\sc Heuristic} and {\sc LinUCB} policies offered a MCQ $12\%$ of the time. In conclusion, in our simulated environment, {\sc LinUCB} noticeably improves student success rate compared to the other two policies.

\section{Experiments}

Following the successful experiments in a simulated environment, we perform a randomized controlled trial on students that have signed up on the platform between December 2021 and February 2022. On sign-up, each student is assigned exercises either by the adaptive {\sc Heuristic} or by the {\sc LinUCB} model. We study $2$ cohorts of students: \textbf{free users}, a diverse set of $44$ students ($21$ under {\sc LinUCB} and $23$ under the heuristic policy) from around the world who signed up on the learning platform for free, and users from a \textbf{customer} organization ($15$ assigned to the {\sc Heuristic} and $11$ to the {\sc LinUCB} policy) using the platform to upskill in data science as part of a broader training program. Students from the second cohort have mandated modules they must finish and tend to be highly motivated regardless of selection policy. Within each cohort, we compare \textit{completion rates}, \textit{skip rates}, and \textit{study time}. Completion and skip rates are local indicators that the exercises we give students are relevant, interesting, and achievable, while study time is a global indicator that the policy is effective at engaging students and motivating them to study on the platform for a longer duration.

\noindent{\bf Completion and Skip Rates}:
As Table \ref{tab:success_table} demonstrates, the students from both cohorts have a substantially higher success rate under {\sc LinUCB} than the adaptive {\sc Heuristic} model. 
We also observe that for both cohorts, students under the {\sc LinUCB} policy have a substantially lower skip rate than under the adaptive heuristic baseline.
These results demonstrate that the {\sc LinUCB} model improves student outcomes and thus does a better job offering more relevant, interesting and achievable exercises to students than the {\sc Heuristic} model.

\begin{table}
\vspace*{-0.6cm}
\centering
    \setlength{\tabcolsep}{0.8em}
    \setlength\extrarowheight{4pt}
    \caption{Exercise outcome rates for various groups of users.}
    \begin{NiceTabular}{|c|c|ccc|}[hlines]
    Cohort & Policy & Skip & Fail & Success\\ 
    \Block{2-1}{Free}
    & {\sc LinUCB} & $\mathbf{7.8} \pm 0.8 \%$ & $4.8 \pm 0.5 \%$ & $\mathbf{87.4} \pm 1.3\%$ \\ 
    & {\sc Heuristic} & $12.5 \pm 1.4 \%$ & $5.2 \pm 0.8 \%$ & $82.4 \pm 1.8\%$ \\ 
    \Block{2-1}{Customer}
    & {\sc LinUCB} & $\mathbf{5.6} \pm 0.4 \%$ & $5.7 \pm 0.4 \%$ & $\mathbf{88.6} \pm 0.9\%$ \\ 
    & {\sc Heuristic} & $8.3 \pm 0.4 \%$ & $ 5.8 \pm 0.3 \%$  & $85.9 \pm 0.7\%$ \\ 
    \end{NiceTabular}
    \label{tab:success_table}
    \vspace*{-0.4cm}
\end{table}

\noindent{\bf Study Time}: Finally, we also observe that students under {\sc LinUCB} across the free cohort spend noticeably more time on the learning platform: the average study time under the adaptive {\sc Heuristic} model is $109$ minutes, and the average study time under the {\sc LinUCB} policy is $174$ minutes. 
For the students in the customer cohort, the average study time under the adaptive {\sc Heuristic} is $265$ minutes, and the average study time under the {\sc LinUCB} policy is $258$ minutes.

\section{Conclusions}

We have provided a framework for developing a model-based reinforcement learning agent based on the LinUCB algorithm, which is capable of both learning from historical student data and online.
This approach outperformed competitive models by achieving significantly higher completion rates, while reducing the rate at which exercises are skipped in two diverse cohorts of students, while also leading to increased study time across cohorts.
These findings demonstrate that the model leads to substantially higher engagement in students.  
In addition, we note that the reinforcement learning model learns autonomously and is expected to improve automatically as more and more students sign up, thus ensuring its scalability and continuous improvement.

In the future, we plan to validate our findings with a larger sample size.
In addition, we will address one of the limitations of the this bandit model -- the requirement that all available exercises pertain to the same topic and that sufficient data is available to reach a point where the bandit can mostly exploit rather than explore. Finally, we plan to explore application of more sophisticated bandit algorithms, such as those that incorporate collaborative filtering. 

\bibliographystyle{splncs04}
\bibliography{mybibliography}

\end{document}